\documentclass[runningheads]{llncs}
\usepackage[T1]{fontenc}
\usepackage{graphicx}
\usepackage{orcidlink}
\usepackage{amsmath}
\usepackage{amssymb}
\usepackage{booktabs}
\usepackage{multirow}
\usepackage{array}
\usepackage{algorithm}
\usepackage{algpseudocode}
\usepackage{algorithmicx}
\usepackage{subcaption}

\begin{document}
\title{Neural Pruning for 3D Scene Reconstruction: Efficient NeRF Acceleration}
\titlerunning{Neural Pruning on NeRF}
%
\author{Tianqi (Kirk) Ding\thanks{These authors contributed equally to this work.}\inst{1}~\orcidlink{0009-0002-9025-0849} \and
Dawei Xiang$^\star$\inst{2}~\orcidlink{0009-0001-2935-0288} \and 
Pablo Rivas\inst{3}~\orcidlink{0000-0002-8690-0987} \and Liang Dong\inst{1}~\orcidlink{0000-0002-8585-1087} }
\authorrunning{T. Ding, D. Xiang, P. Rivas, and L. Dong}
%
\institute{Department of Electrical and Computer Engineering, Baylor University, USA
\email{\{Kirk\_Ding1,Liang\_Dong\}@Baylor.edu} \and 
Department of Computer Science and Engineering, University of Connecticut, USA
\email{ieb24002@uconn.edu} \and 
Department of Computer Science, Baylor University, USA
\email{Pablo\_Rivas@Baylor.edu}}
\maketitle              
\begin{abstract}
Neural Radiance Fields (NeRF) have become a popular 3D reconstruction approach in recent years. While they produce high-quality results, they also demand lengthy training times, often spanning days. This paper studies neural pruning as a strategy to address these concerns. We compare pruning approaches, including uniform sampling, importance-based methods, and coreset-based techniques, to reduce the model size and speed up training. Our findings show that coreset-driven pruning can achieve a 50\% reduction in model size and a 35\% speedup in training, with only a slight decrease in accuracy. These results suggest that pruning can be an effective method for improving the efficiency of NeRF models in resource-limited settings.

\keywords{NeRF \and Model Compression \and Neural Pruning \and 3D Reconstruction \and Efficiency}
\end{abstract}

\section{Introduction}

Neural Radiance Fields (NeRF) have become an influential approach for synthesizing 3D scenes from 2D images while preserving detailed visual quality. Despite their effectiveness, they often demand extensive computational resources, requiring long training times that can extend over multiple days \cite{tianqi2025nerfbaseddefectdetection}. Researchers have explored strategies to address these concerns, primarily by compressing the model or pruning less important parameters. Such methods seek to reduce computational overhead without severely affecting the ability to represent scene details. Existing work underscores the importance of this endeavor: long-running NeRF processes may impede real-world tasks like cultural heritage documentation and architectural modeling, where rapid generation of 3D content is vital~\cite{10.5194/isprs-archives-xlviii-m-2-2023-1051-2023,10.1117/12.3031605}.

One compression strategy uses uniform sampling. By evenly distributing sample points, this method decreases computational load in a straightforward manner. However, uniform sampling may cause a loss of detail in regions where small features matter \cite{10.48550/arxiv.2111.15552,10.48550/arxiv.2106.05264}, and it may not suffice for applications that require precise rendering of scene intricacies \cite{10.1109/icce56470.2023.10043459}. Another approach is importance-based sampling, which allocates more resources to parts of the input space that have a larger impact on the final image. In doing so, it can enhance training efficiency and preserve fine-grained features \cite{10.48550/arxiv.2111.15552}.

A further extension involves coreset methods, in which smaller, representative subsets of the data are selected to approximate the full dataset. This approach aligns with the ultimate goal of model compression by retaining essential information while discarding details that have negligible influence on output quality. Studies on coreset-driven pruning have shown promising efficiency improvements, suggesting that such techniques can maintain high-fidelity novel views while using fewer samples \cite{10.48550/arxiv.2111.15552,10.1609/aaai.v37i3.25469,10.1109/cvpr46437.2021.01018}. Compression ideas, including voxel-based representations, have also surfaced in other works such as TinyNeRF, which achieved significant size reductions \cite{10.1609/aaai.v37i3.25469}. While these methods vary in details, they share a common objective of conserving computational resources without undermining visual realism.

As additional pruning and compression methods gain traction in 3D reconstruction, evaluations reveal key trade-offs. Some strategies greatly shorten training cycles but may sacrifice the ability to represent subtle visual information~\cite{10.1109/iccv48922.2021.00583,10.48550/arxiv.2111.15552}. Balancing compression and performance is crucial in domains that require both rapid 3D synthesis and detailed rendering. Although there have been efforts to integrate features of importance-based sampling with coreset ideas, there remains a pressing need for principled evaluations of these hybrid solutions \cite{10.3390/app131810174,10.48550/arxiv.2105.06466}.

This paper aims to systematically investigate pruning-based compression methods for NeRF. By comparing uniform, importance-based, and coreset approaches, we highlight how each one affects both efficiency and reconstruction fidelity. Our results indicate that it is possible to compress NeRF models substantially while preserving most of their scene representation capabilities. We believe that this study not only advances the understanding of pruning in NeRF but also informs broader research on compressing 3D reconstruction architectures \cite{10.48550/arxiv.2012.08503,10.1109/cvpr52688.2022.01807}. The contributions of this paper are summarized as follows:
\begin{itemize}
    \item Demonstrate the feasibility of neural pruning techniques on 3D Reconstruction networks.
    \item Achieve decent performance on compressing model size and accelerating computation speed.
\end{itemize}

The rest of the paper is organized as follows. Section 2 reviews related work on pruning-based strategies in NeRF and related 3D reconstruction models. Section 3 introduces the methodology and experimental setup. Section 4 discusses our results in detail, including analysis of accuracy and speed gains. Finally, Section 5 presents conclusions and future research directions.

\section{Related work}

\subsection{Neural Network Pruning} Neural network pruning seeks to cut down unnecessary parameters—either connections or neurons—in a trained model, with the goal of reducing computational demands and memory usage while keeping acceptable performance. Common pruning approaches fall into two categories: unstructured and structured.

\paragraph{Unstructured Pruning}
Unstructured pruning eliminates individual weights in the network, typically those whose magnitudes are below a chosen threshold. Although this can dramatically reduce the parameter count, it also creates irregular memory patterns, which can hinder efficient implementation on hardware designed for dense operations. Our initial experimentation with “Edge Pruning,” where the smallest-magnitude connections are removed, belongs to this category.

\paragraph{Structured Pruning}
Structured pruning, by contrast, removes entire neurons, filters, or even layers. This approach focuses on pruning higher-level architectural components, resulting in networks that maintain more regular structures suitable for hardware acceleration. However, removing significant blocks of the network can lead to greater performance drops if done too aggressively. In our later experiments, structured pruning became the main approach for boosting NeRF’s training speed because it more naturally reduces the size of the fully connected layers.

\subsection{Pruning Methods}

\paragraph{Uniform Sampling} Uniform sampling selects neurons from a given layer at equal probability, thereby reducing network size in a straightforward manner. This random strategy offers computational simplicity since it does not require additional metrics or calculations to determine which neurons to remove. However, because it does not account for the contribution that individual neurons make to network performance, its outcomes can be suboptimal. In practice, uniform sampling often removes valuable neurons alongside those that are genuinely redundant, leading to potential decreases in overall accuracy.

\paragraph{Importance Pruning} Importance pruning targets neurons or connections deemed least critical to the network's inference. This entails evaluating metrics such as weight magnitudes or estimated contributions to the network's output, allowing for the selective removal of parameters while striving to maintain predictive capability. By eliminating less impactful elements, the model can become more efficient without a large decrease in quality. Importance pruning has proven effective under various settings, particularly when datasets are evenly distributed or when memory resources are restricted. For instance, \cite{hewahi2019neural} demonstrates how this technique can streamline network architectures while preserving performance across different data domains.

\paragraph{Coreset} Coreset-based pruning constructs a smaller yet representative subset of points to approximate the original layer or dataset. Under this method, each candidate neuron is assigned a weight, and sampling probabilities are determined in proportion to these weights. As a result, the chosen subset reflects the most informative components of the network, reducing computational costs while retaining critical features. Section 3.6 will provide an algorithm that further illustrates the implementation details.

In the work of \cite{mussay2021data}, neurons were viewed as coreset elements, with the methodology selectively removing those deemed redundant in a bottom-up manner, thus preserving accuracy while recalibrating remaining connections. Similarly,~\cite{dubey2018coreset} introduced a coreset-focused scheme for compressing convolutional neural networks, exploiting internal redundancies to reduce storage footprints and expedite inference, all without additional retraining steps.

\subsection{Neural Radiance Fields (NeRF)}
\begin{figure}[h!]
    \centering
    \includegraphics[width=\textwidth]{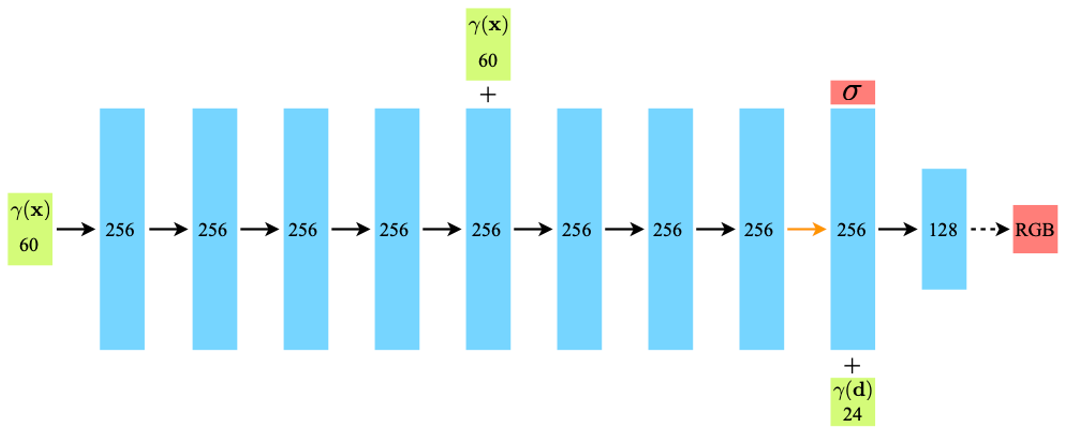}
    \caption{A visualization of the MLP architecture from \cite{mildenhall2021nerf}. The network accepts $\gamma(x)$, a positional encoding of the 3D coordinates, along with $\gamma(d)$, an encoding of the viewing direction. It then outputs the density $\sigma$ and RGB color of the point $(x,y,z)$.}
    \label{fig:nerf_mlp}
\end{figure}

Neural Radiance Fields (NeRF), first introduced by \cite{mildenhall2021nerf}, have been widely adopted for producing high-fidelity 3D scene reconstructions from 2D image sets. At its core, NeRF relies on a multilayer perceptron (MLP) to map each 3D coordinate $(x,y,z)$ and viewing direction $(\theta,\phi)$ to a density value $\sigma$ and a corresponding RGB color. By accumulating these densities and colors along any desired viewing ray, NeRF renders photorealistic images from novel perspectives. Fig.~\ref{fig:nerf_mlp} illustrates the MLP architecture used in NeRF.

\paragraph{Model compression on NeRF} Efforts to compress or prune NeRF have been relatively limited. For instance, \cite{isik2021neural} explores edge-level pruning to highlight sparsity in the network’s weights, whereas \cite{tang2022compressible} leverages singular value decomposition (SVD) to yield lower-rank approximations of the NeRF representation. Additionally, \cite{xie2023hollownerf} proposes pruning hash tables for 3D mesh-based reconstructions in Instant-NGP. Despite these advancements, the application of neuron-level pruning and coreset-based methods to NeRF remains underexplored, offering a potential avenue for more efficient 3D reconstructions without sacrificing realism.

\paragraph{}
In addition to prior studies on pruning and efficient NeRF representations, recent developments in parameter-efficient learning provide complementary insights. For example, VMT-Adapter \cite{xin2024vmt} and MmAP \cite{xin2024mmap} introduce efficient transfer learning methods for multi-task vision, which share structural similarities with importance-based pruning in our framework. Moreover, V-PETL \cite{xin2024vpetl} and the survey on parameter-efficient fine-tuning \cite{xin2024parameter} highlight practical strategies to maintain model performance with fewer parameters, echoing the goals of our pruning techniques. These ideas inspire how pruning can be embedded into broader visual systems while retaining generalization. Additionally, reinforcement learning-based recognition of unordered robotic targets \cite{mao2025reinforcement} demonstrates the importance of selecting representative visual features under constrained resources, which aligns with our coreset sampling approach.

\section{Methods and Experiments }

\subsection{Sparcity in NeRF MLP}
Before trying to compress the MLP in NeRF, we need to prove that there is indeed sparsity in NeRF model. Therefore we first did the research on the neural representation of NeRF. 

\begin{figure}[h!]
    \centering
    \includegraphics[width=0.75\textwidth]{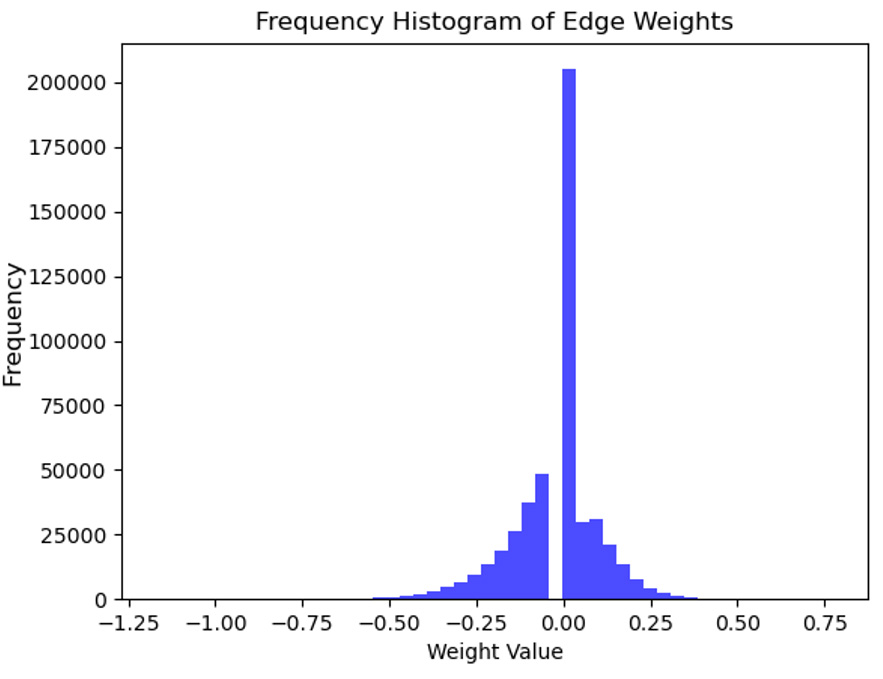}
    \caption{Frequency Histogram of edge weights in NeRF. A large percent of edge weights fall into the range $[0, 0.05]$. }
    \label{fig:nerf_freq}
\end{figure}

We have extracted the edge weights in the above MLP model. We first drew a frequency distribution Fig.~\ref{fig:nerf_freq}. And we found that a large amount of edge weights are indeed very small (less than 0.05). This indicates that the MLP in NeRF is indeed sparse, and there is potential space for compression.

\subsection{Pruning the edges }
Based on the discovery of the sparsity, we then tried to prune some low-weight edges: we discard the edges that have a low weight.  When threshold is 0.05, only 40\% edges will be kept and 60 \% will be discarded. The edge pruning result is shown on Table \ref{tab:edge_prune}:
\begin{table}[h!]
    \centering
    \caption{Edge pruning comparison. 
    PSNR denotes the peak signal-to-noise ratio, 
    where higher values indicate closer similarity to ground truth.}
    \label{tab:edge_prune}
    \begin{tabular}{@{}lccc@{}}
        \toprule
        & \textbf{Threshold = 0} & \textbf{Threshold = 0.05} & \textbf{Threshold = 0.1} \\ 
        \midrule
        \textbf{Remaining Percentage} & 100\% & 40\% & 20\% \\
        \textbf{PSNR (Test Set)}      & 21.5  & 21.3 & 20.8 \\
        \bottomrule
    \end{tabular}
\end{table}

The trends of increasing the remaining percentage and corresponding power-to-signal noise ratio (PSNR) indicate that eliminating redundant edges has minimal impact on the outcome. Hence, we will shift our focus from edge pruning to neuron pruning. The PSNR is defined as:
\begin{equation}
    PSNR = 10 \log_{10} \left(\frac{MAX_I^2}{MSE} \right) , 
\end{equation}
where $MAX_I$ is the maximum possible pixel value of the image, and Mean Squared Error (MSE) is defined as:
\begin{equation}
    MSE = \frac{1}{m n} \sum_{i=0}^{m-1} \sum_{j=0}^{n-1} [I(i,j) - K(i,j)]^2 , 
\end{equation}
where $I(i,j)$ is the original image, $K(i,j)$ is the reconstructed image, and $m \times n$ are the dimensions of the images.

\subsection{Problem of Unstructured Edge Pruning}
While pruning edges can remove many low-magnitude connections, it does not reduce the overall size of the layer weights. Consequently, the training time remains largely unaffected, because the weight matrices between layers still have the same dimensions. To address this, we explored pruning at the neuron level. In particular, the original model utilizes seven fully connected layers, each sized \(256 \times 256\), with one additional layer for incorporating the viewing direction. By compressing these to \(64 \times 64\), the network's parameter count could theoretically be reduced to one-sixteenth of its initial size.

\subsection{Uniform Sampling}
A straightforward way to prune neurons is to select them at random. In this approach, each neuron has an equal probability of being retained, and the model is subsequently retrained. Table~\ref{tab:uni} illustrates the impact of uniform sampling, where we reduce each layer from \(256 \times 256\) neurons to \(64 \times 64\). Although this technique does shrink the parameter count, it markedly degrades the peak signal-to-noise ratio (PSNR), indicating that random pruning can remove many crucial neurons.

\begin{table}[ht]
    \centering
    \caption{Performance with uniform sampling. 
    We randomly select 64 neurons (out of 256) in each layer, then retrain the model.}
    \label{tab:uni}
    \begin{tabular}{@{}lcc@{}}
        \toprule
        & \textbf{Baseline} & \textbf{Uniform Sampling} \\
        \midrule
        Connection Layer Size & \(256 \times 256\) & \(64 \times 64\) \\
        PSNR                  & 21.5              & 16.5           \\
        Model Size            & 2.38\,MB          & 0.7\,MB        \\
        \bottomrule
    \end{tabular}
\end{table}

The substantial drop in performance highlights two points: \emph{(i)} random selection of neurons is ineffective at preserving important features, and \emph{(ii)} a more informed pruning criterion is necessary to maintain quality while reducing model size.

\subsection{Importance Pruning}
The results in Table~\ref{tab:uni} show that randomly discarding neurons degrades performance, underscoring the need for a more selective pruning strategy. To address this, we compute an importance score for each neuron and then discard those with the lowest scores.

\begin{figure}[h!]
    \centering
    \includegraphics[width=0.6\textwidth]{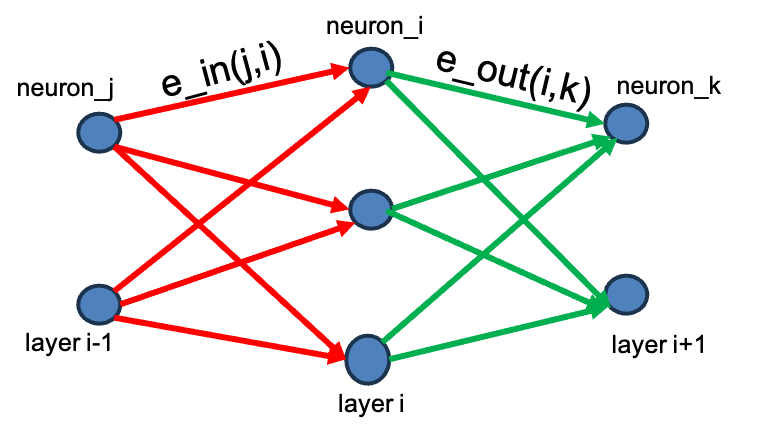}
    \caption{An illustration of the importance weight calculation. Red arrows represent incoming edges for neurons in layer \(i\), 
    whereas green arrows indicate outgoing edges.}
    \label{fig:nerf_cal_weight}
\end{figure}

A neuron’s importance can be derived from its incoming and outgoing edge weights. Let the neuron \(i\) belong to layer \(L_{k}\), with its preceding and subsequent layers denoted by \(L_{k-1}\) and \(L_{k+1}\), respectively. We define:
\begin{equation}
    w_{\text{in}}(i) = \frac{\sum_{j \in L_{k-1}} \left|e_{ji}\right|}{\left|L_{k-1}\right|} ,
    \label{eq:win}
\end{equation}
\begin{equation}
    w_{\text{out}}(i) = \frac{\sum_{r \in L_{k+1}} \left|e_{ir}\right|}{\left|L_{k+1}\right|} ,
    \label{eq:wout}
\end{equation}
where \(e_{ji}\) and \(e_{ir}\) represent the connection weights from neuron \(j\) to \(i\), and from \(i\) to \(r\), respectively. Fig.~\ref{fig:nerf_cal_weight} illustrates this concept, showing incoming and outgoing connections for a target neuron in layer \(i\).

Using these definitions, we prune the neurons with the lowest importance scores and then retrain the resulting network. Table~\ref{tab:importance} presents the outcomes when using \(w_{\text{in}}\), \(w_{\text{out}}\), or their product as the importance criterion.

\begin{table}[ht]
    \centering
    \caption{Results of different importance metrics.
    We compare pruning by \(w_{\text{in}}\), \(w_{\text{out}}\), 
    and their product, followed by retraining.}
    \label{tab:importance}
    \begin{tabular}{@{}llr@{}}
        \toprule
        & \textbf{Method} & \textbf{PSNR} \\
        \midrule
        No Pruning & Baseline & 21.5 \\
        \midrule
        \multirow{3}{*}{Importance Pruning} 
          & Select Neurons with \(w_{\text{in}}\) & 19.5 \\
          & Select Neurons with \(w_{\text{out}}\) & 20.0 \\
          & Select Neurons with \(w_{\text{in}} \times w_{\text{out}}\) & 20.0 \\
        \bottomrule
    \end{tabular}
\end{table}

Notably, pruning based on \(w_{\text{out}}\) yields slightly higher PSNR values compared to using \(w_{\text{in}}\). This finding suggests that focusing on how each neuron influences subsequent layers can be more effective in preserving overall model performance.

\subsection{Coreset}
Encouraged by the performance of importance pruning, we evaluated a coreset-based approach \cite{mussay2019data} to select a representative subset of neurons in each layer. This technique assigns a sampling probability to each neuron, then draws a smaller set of neurons that collectively preserve most of the model’s representational capacity. In practice, we follow a procedure similar to that of \cite{mussay2019data}, using the product of \( w_{\text{in}}(i) \) and the average outcoming weight \( w_{\text{out}}(i) \) to guide neuron selection. For simplicity, we replace the maximum outcoming edge weight with its average counterpart. The general pseudocode is shown in Algorithm \ref{coreset_algo}.

\begin{algorithm}[h!]
\caption{Coreset}
\begin{algorithmic}[1]
\Statex \textbf{Input:} 
\Statex \hspace{\algorithmicindent}   $w_{in}(i)$ and $w_{out}(i)$ for all the neurons in the layer $k$ with $i \in \{1 ,\ldots, |L_k|\}$ where $|L_k|$ is the total number of neurons in layer $k$, 256, in our case.
\Statex \hspace{\algorithmicindent}  An integer (sample size) $m \geq 1$, 64, in our case.
\Statex \hspace{\algorithmicindent}  An (activation) function $\varphi : \mathbb{R} \rightarrow [0, \infty)$, ReLU(), in our case.
\Statex \hspace{\algorithmicindent}  An upper bound controller $\beta > 0$, 3, in our case.
\Statex \textbf{Output:} 
\Statex \hspace{\algorithmicindent} A weighted set $(C, u)$ which corresponding to the coreset of the layer; 
\ForAll{$i \in \{1 ,\ldots, |L_k|\} $}
    \State $\displaystyle pr(i) := \frac{ w_{in}(i) \varphi(\beta* w_{out}(i))}{\sum_{j \in L_k} w_{in}(j) \varphi(\beta * w_{out}(j)) }$

    \State $u(i) := 0$
\EndFor
\For{$m$ iterations}
    \State Sample a point $q$ from $L_k$ with probability $pr(q)$.
    \State $C := C \cup \{q\}$
    \ForAll{$i \in [L_k]$}
        \State $\displaystyle u_i(q) := u_i(q) + \frac{w_{out}(q)}{m \cdot pr(q)}$
    \EndFor
\EndFor
\State \Return $(C,u_1,\dots,u_k)$
\end{algorithmic}
\label{coreset_algo}
\end{algorithm}

After selecting neurons based on these coreset probabilities, we retrain the compressed network. Table~\ref{tab:final_cmpr} summarizes the outcomes of pruning each \(256 \times 256\) fully connected layer down to either \(128 \times 128\) or \(64 \times 64\).

\begin{table}[h!]
    \centering
    \caption{Performance comparison using coreset sampling. Compressing from \(256 \times 256\) to \(128 \times 128\) preserves the highest PSNR 
    while noticeably reducing both training time and model size.}
    \label{tab:final_cmpr}
    \begin{tabular}{@{}lccc@{}}
    \toprule
    \textbf{Connection Layer Size} & \(256 \times 256\) & \(128 \times 128\) & \(64 \times 64\) \\
    \midrule
    Model Parameters  & 595K     & 288K      & 177K \\
    Model Size        & 2.38\,MB & 1.14\,MB  & 0.7\,MB \\
    PSNR              & 21.5     & \textbf{21.3} & 20.1 \\
    Training Time 
    (min / 100k iterations) & 78.75 & \textbf{51.25} & 46.25 \\
    \bottomrule
    \end{tabular}
\end{table}

Notably, reducing the layer dimensions from \(256 \times 256\) to \(128 \times 128\) cuts the training time by roughly 35\% and halves the model size, while maintaining a PSNR of 21.3. This highlights the viability and effectiveness of neuron pruning in NeRF, particularly via coreset-based sampling.

\begin{figure}[h!]
    \centering
    \begin{subfigure}{0.45\textwidth}
        \centering
        \includegraphics[width=\linewidth]{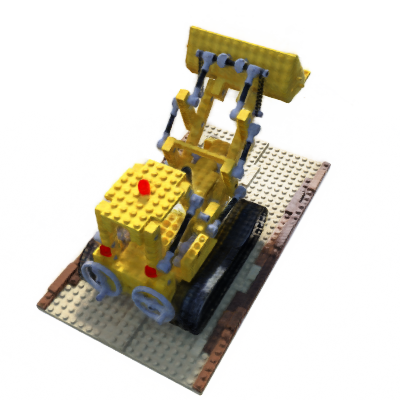}
        \caption{Baseline: $256\times 256$, PSNR: 21.5}
        \label{fig:256}
    \end{subfigure}
    \hfill
    \begin{subfigure}{0.45\textwidth}
        \centering
        \includegraphics[width=\linewidth]{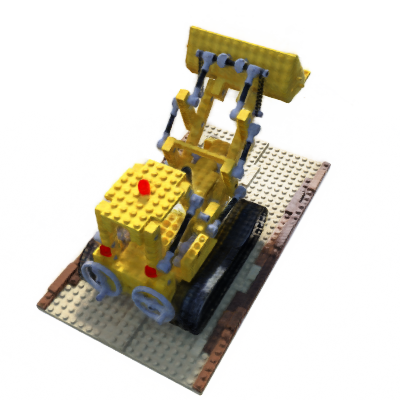}
        \caption{Coreset: $128\times 128$, PSNR: 21.3}
        \label{fig:128}
    \end{subfigure}
    \hfill
    \begin{subfigure}{0.45\textwidth}
        \centering
        \includegraphics[width=\linewidth]{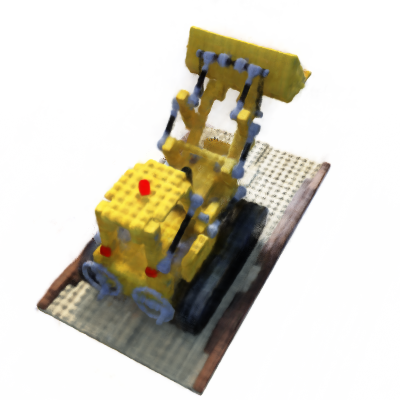}
        \caption{Coreset: $64\times 64$, PSNR: 20.1}
        \label{fig:64}
    \end{subfigure}
    \caption{Performance comparison visualization for different compression scales.}
    \label{fig:performance-comparison}
\end{figure}

To visualize how pruning affects output quality, Fig.~\ref{fig:performance-comparison} illustrates sample renderings at different compression scales. As shown in the figure, the perceptual differences among the various compression scales are not dramatic, even though the numerical metrics do shift. The baseline model (\(256 \times 256\)) produces the highest PSNR of 21.5, but compressing the layers to \(128 \times 128\) only slightly degrades performance (PSNR of 21.3), while providing substantial reductions in both training time and model size. By contrast, the more aggressive pruning to \(64 \times 64\) retains most of the scene’s overall structure yet reveals small losses of detail (PSNR of 20.1). These visual comparisons corroborate our quantitative findings and highlight that coreset pruning can achieve a good balance between efficiency and scene fidelity.

\section{Conclusion}
In this paper, we compared multiple neural pruning strategies, including uniform sampling, importance pruning, and coreset-based methods, for compressing NeRF’s MLP layers. Our experiments showed that random approaches often degrade quality significantly, while importance-based pruning produces better results by targeting less influential neurons. Among the tested techniques, coreset pruning provides a balanced outcome, achieving a significant reduction in both model size and training time, with only a slight performance loss in terms of peak signal-to-noise ratio.

These findings highlight that neuron-level pruning is an effective way to accelerate NeRF training, taking advantage of latent sparsity within the network. Future research may also draw from parameter-efficient transfer learning benchmarks such as V-PETL \cite{xin2024vpetl} or causal representation learning like CausalSR \cite{lu2025causalsr}, to integrate pruning with causal inference or visual generalization strategies. These directions may enable NeRF acceleration not only in synthetic benchmarks but also in more dynamic or interactive 3D environments. We expect that insights from this research will inform new developments in network compression for a broad range of 3D reconstruction and rendering applications.

\begin{credits}
\subsubsection{\ackname} Part of this work was funded by the National Science Foundation under grants CNS-2210091, OPP-2146068, CHE-1905043, and CNS-2136961; and by the Department of Education under grant P116Z230151.

\subsubsection{\discintname}
The authors have no competing interests to declare that are
relevant to the content of this article.
\end{credits}
 
%
%
%

%

\end{document}